\documentclass{bmvc2k}

\usepackage{amsmath,amsfonts,bm}

\def\eqref#1{equation~\ref{#1}}

\def\1{\bm{1}}

\DeclareMathAlphabet{\mathsfit}{\encodingdefault}{\sfdefault}{m}{sl}
\SetMathAlphabet{\mathsfit}{bold}{\encodingdefault}{\sfdefault}{bx}{n}

\usepackage{algorithm}
\usepackage{algorithmic,booktabs,adjustbox,hhline, colortbl,tabularx}
\usepackage{multirow,makecell}
\usepackage[symbol]{footmisc}
\usepackage{xcolor}
\definecolor{amber}{RGB}{214,120,0}

\usepackage{algorithmic}
\usepackage{pifont}
\usepackage[accsupp]{axessibility}
\usepackage{pifont}
\usepackage[accsupp]{axessibility}
\newcommand{\etal}{\textit{et al.}}
\usepackage{xcolor}
\usepackage{booktabs}
\usepackage{multirow}
\newcommand{\cmark}{\textcolor{blue}{\ding{51}}}
\newcommand{\xmark}{\textcolor{red}{\ding{55}}}
\definecolor{navyblue}{RGB}{0,0,128}

\title{CoCam4D: Geometry-Aware Cooperative 4D Perception for Camera-Only Autonomous Driving}

\runninghead{Pahari \etal}{cocam4d}

\addauthor{Soham Pahari$^{\mbox{\sffamily\scriptsize 1,}}$}
{soham.109424@stu.upes.ac.in}{3}

\addauthor{Sudip Das\textsuperscript{*}}
{sudip.das@valeo.com}{2}

\addauthor{Arindam Das}
{arindam.das@valeo.com}{2}

\addauthor{Ujjwal Bhattacharya}
{ujjwal@isical.ac.in}{3}

\addinstitution{
University of Petroleum and Energy Studies, Dehradun
}

\addinstitution{
anSWer India, Valeo Brain
}

\addinstitution{
Indian Statistical Institute, Kolkata
}

\def\etal{\emph{et al}\bmvaOneDot}

\begin{document}

\maketitle


\begin{abstract}

Autonomous vehicles often suffer from limited perception due to occlusions, blind spots, limited sensor range and complex nature of the surroundings. 
Multi-agent collaborative perception (CP) addresses these challenges by allowing vehicles to share sensory information seamlessly and reconstruct the scene. 
However, camera-only perception remains fundamentally restricted due to uncertain nature of the distance-dependent monocular depth.
We propose \textbf{CoCam4D}, a Bayesian framework for collaborative perception that explicitly models geometric uncertainty. 
It uses a VGGT-based feedforward network to generate 3D Gaussian scene representations with uncertainty estimates, helping multiple vehicles or agents to combine their \textit{observations} efficiently.
By sharing compact Gaussian primitives, reliable observations from one agent reduces the limitation of depth uncertainty in another without requiring LiDAR sensor data.  
To support real-world deployment, we introduce Dynamic Object Primitives (DOPs), a compact 35-byte representation for efficient C-V2X communication.
Extensive experiments show that our proposed metho consistently outperforms recent vision-only methods, achieving improvements of 11.48\% on OPV2V+ and 10.62\% on DAIR-V2X-C, suggesting a promising geometrically grounded direction toward LiDAR-free autonomous driving.


\end{abstract}

\vspace{-15pt}
\section{Introduction}

With the rapid advancement of connected and autonomous driving technologies, multi-agent/vehicle collaborative perception (CP) has become a promising paradigm for intelligent transportation systems.
Through Vehicle-to-Everything (V2X) communication, connected agents exchange perceptual information to cooperatively understand complex driving environments beyond the capability of a single vehicle.
This shared perception helps to build complete and reliable understanding of traffic scenes, leading to significant improvements in safety, mobility, and overall transportation efficiency~\cite{cp_survey}.
Early CP studies primarily focused on traditional single-agent tasks such as 3D object detection~\cite{who2com,where2comm,cppc,ermvp,v2xvit} and 
semantic segmentation~\cite{v2vnet,disconet}, 
typically by fusing latent BEV features. 
Although LiDAR point clouds have long been used for their precise geometric measurements, recent advances in camera-based 3DOD, such as CoCa3D ~\cite{coca3d}, have demonstrated competitive performance,  highlighting cameras as viable, cost-effective sensors for complex 3D scene understanding.
However, existing camera-based cooperative methods usually share compressed BEV feature grids~\cite{v2vnet,cobevt,where2comm,disconet} without modeling uncertainty. 
As a result, fusion across vehicles relies on learned attention, without optimality guarantees and failing to distinguish between more reliable near-range observations and less certain far-range observations.


\begin{figure}[t] 
    \centering
    \includegraphics[width=\textwidth]{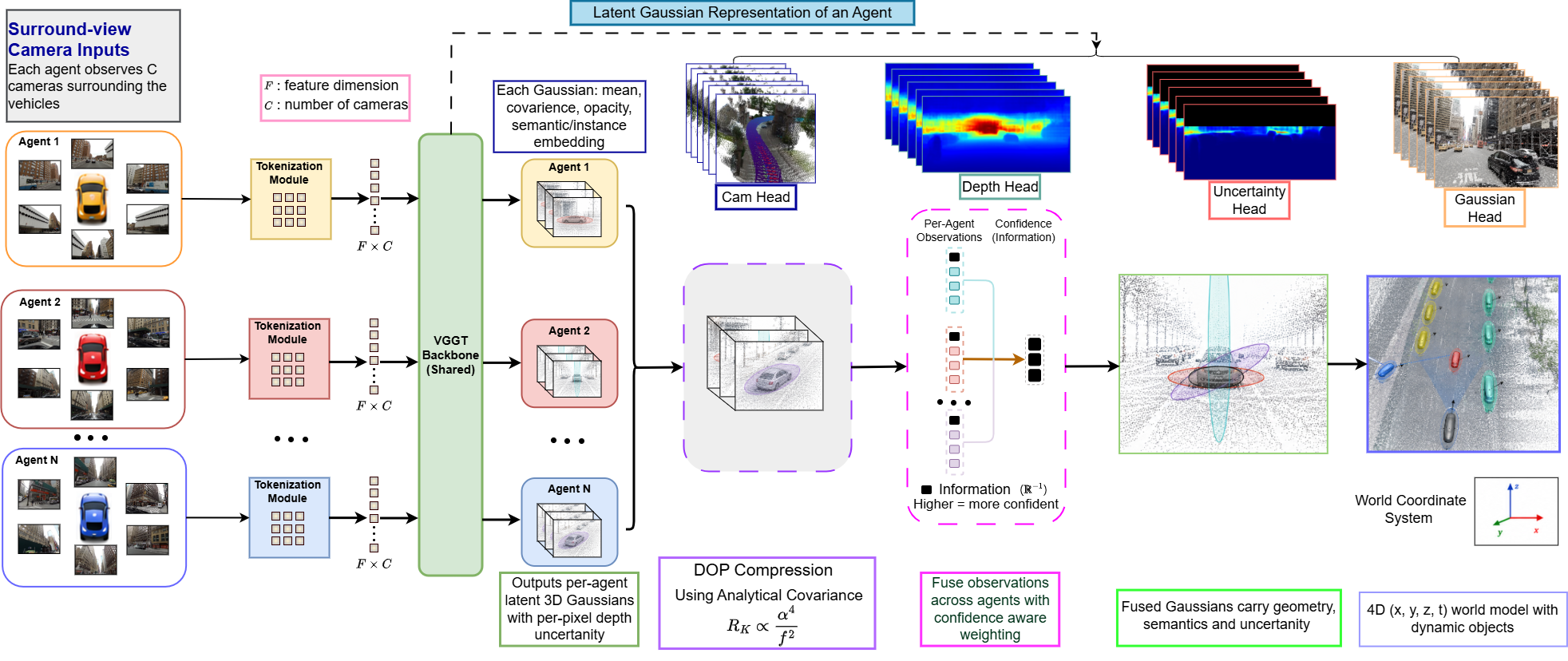}
    \vspace{-5mm}
    \caption{\textbf{CoCam4D overview.} Each CAV tokenises its $F{\times}C$ surround-view frames and passes them through a shared-weight VGGT encoder to produce velocity-aware 3D Gaussians with per-pixel depth uncertainty in a single forward pass---no LiDAR or per-scene optimisation required. Dynamic objects are compressed into \emph{Dynamic Object Primitives} (DOPs, ${\approx}35$ bytes/object) encoding state and analytically derived covariance $\mathbf{R}_k\!\propto\!z^4/f^2$, then fused across vehicles via an information filter $\mathbf{Y}_\text{fused}{=}\sum_j\mathbf{R}_j^{-1}$ that automatically assigns weights to the confident lateral views over uncertain head-on observations, yielding a cooperative 4D world model that surpasses single-vehicle LiDAR at $N{=}5$ agents.}
\label{fig:overview} 
\vspace{-5mm}
\end{figure}

The core limitation of camera-only autonomous driving arises from the issues of perspective geometry, where monocular depth follows 
$
z \propto \frac{f \cdot H}{h},
$ where $z$ is the object depth, $f$ is the camera focal length, $H$ is the real-world object height, and $h$ is the object height in the image.
The associated depth uncertainty increases quadratically with object depth,
$
\sigma_z \propto \frac{z^2}{f},
$ where $\sigma_{z}$ represents the depth uncertainty,
making long-range perception fundamentally unreliable~\cite{lu2021geometry,zhang2021objects,das2024mmprune4u,hazra2025reflective}. 
Existing single-vehicle vision approaches~\cite{bevformer,bevdepth,streampetr} do not fully overcome this ambiguity, as it originates from the physics of perspective projection rather than the model capacity.  
CP reduces this shortcoming by using observations from multiple views. When several vehicles observe the same object from different angles, the combined views reduce depth uncertainty and improve 3D localization~\cite{coca3d}.
Yet, principled Bayesian fusion for multi-vehicle camera-only perception remains largely unexplored.  
Recent works~\cite{3dgs,edgs} inspired by 3D Gaussian Splatting use direction-aware Gaussian representations that remain stable under rigid transformations. 
These representations are compact and efficient for communication in cooperative scene understanding~\cite{chen2026vision}.
However, existing methods still rely on black-box fusion and lack explicit modeling of the distance-dependent, anisotropic uncertainty induced by perspective projection. 
Motivated by predictive coding theories \cite{denham2020predictive} of perception, we introduce a probabilistic collaborative perception framework with explicit uncertainty propagation and Bayesian fusion.

In this work, we propose \textbf{CoCam4D}, the first feedforward cooperative 4D perception framework that reconstructs velocity-aware 3D Gaussian scenes per vehicle and fuses them across agents via a principled information-filter-based Gaussian product\cite{williams2001products}.
Each vehicle performs a single forward pass through a VGGT-based backbone~\cite{vggt} with causal masked attention, predicting per-pixel metric depth, 3D Gaussians with lifespan, per-pixel velocities, and depth uncertainty, without requiring LiDAR or additional cross-vehicle calibration refinement beyond the dataset-provided intrinsics and extrinsics used in CoCa3D~\cite{coca3d}.
Dynamic objects are encoded as compact Dynamic Object Primitives (DOPs) and transmitted at approximately 35 bytes per object via V2V communication. 
Cross-vehicle fusion combines information from multiple vehicles using confidence-based weights derived from depth uncertainty.
This creates a shared world model that is complete and reliable than the view from a single vehicle.

Contributions of this study are summarized as follows:

\begin{itemize}
    \item \textbf{CoCam4D}, a Vision only  cooperative 4D perception framework
    that reconstructs and fuses velocity-aware Gaussian scenes across vehicles
    in a single forward pass.
    \vspace{-8pt}
    \item An end-to-end uncertainty chain from per-pixel depth error to
    Gaussian covariance and information-filter weighting, allowing uncertainty-driven cooperative fusion.
    \vspace{-8pt}
    \item A compact Dynamic Object Primitive (DOP) fitting one C-V2X packet,
    preserving full object state and observation covariance for real-time V2V
    communication.
    \vspace{-8pt}
    \item Through a cooperative density study, we show that five cooperating camera agents achieve performance comparable to a single-vehicle LiDAR system.
\end{itemize}

\vspace{-23pt}

\section{Related Work}
Existing works in this area is categorized into five primary streams: multi-agent perception, camera-based 3D scene understanding, cooperative perception, feedforward 4D reconstruction, and 3D Gaussian representations for driving and collaborative systems
\vspace{-5mm}
\paragraph{Camera-Only 3D Perception.}

Camera-only 3D object detection has advanced quickly, with bird’s-eye-view (BEV) representations helping to combine information from multiple views in a unified way.
BEVFormer~\cite{bevformer} and Reflective Teacher \cite{hazra2025reflective} introduce spatio-temporal attention for BEV aggregation, while BEVDepth~\cite{bevdepth} improves geometric fidelity via explicit depth supervision. 
StreamPETR~\cite{streampetr} models temporal information using persistent object-centric queries and achieves strong performance on the nuScenes Dataset~\cite{caesar2020nuscenes}. 
HeightFormer~\cite{wu2024heightformer} introduces explicit height modeling in BEV without additional supervision which reduces depth ambiguity by decoupling height estimation from projection.
IAE-BEV~\cite{wang2025iae} mitigates feature degradation in LSS pipelines via instance-adaptive enhancement with object-aware pooling and attention. 
DenseBEV~\cite{dahling2026densebev} further reformulates BEV grids as dense object queries for structured feature-to-prediction mapping.
Despite these advances, monocular geometry remains fundamentally limited since depth estimation relies on perspective cues with uncertainty growing as $\sigma_z \propto z^2 / f$~\cite{lu2021geometry, zhang2021objects}, degrading long-range perception. Cooperative perception addresses this by sharing information across agents. 
\vspace{-10pt}
\paragraph{Cooperative Perception.}

Connected autonomous agents share perception through Vehicle-to-Everything (V2X) communication, which helps reduce problems like occlusion, limited field of view, and sparse observations~\cite{cp_survey}.
Methods are commonly categorized into early, intermediate, and late fusion depending on the exchange stage. 
Early fusion transmits raw sensor data~\cite{vieye}, which keeps full geometric details but requires high communication cost.
Late fusion shares object-level outputs~\cite{vips,otvic}, which reduces bandwidth but loses some spatial detail.
Intermediate fusion has thus become dominant, where compact latent features are shared among agents~\cite{v2vnet,where2comm,cobevt,v2xvit,v2xvitv2,cppc}.  
F-Cooper~\cite{chen2019fcooper} introduced BEV feature fusion, followed by graph-based V2VNet~\cite{v2vnet}, adaptive Where2comm~\cite{where2comm}, and transformer-based V2X-ViT~\cite{xu2022v2xvit} and CoBEVT~\cite{xu2022cobevt}, establishing intermediate fusion as the standard for bandwidth-efficient CP.  
Large-scale benchmarks such as OPV2V~\cite{xu2022opencood}, V2X-Sim~\cite{li2022v2xsim}, and V2XSet~\cite{xu2022v2xvit}, along with real-world datasets including DAIR-V2X~\cite{yu2022dair}, V2X-Real~\cite{xiang2024v2x}, and TUMTraf-V2X~\cite{zimmer2024tumtraf}, support evaluation under realistic sensing and communication conditions.  

Most CP methods remain LiDAR-centric. Camera-only cooperative perception is less explored due to monocular depth ambiguity and multi-view misalignment.
Existing vision-based approaches typically exchange deterministic BEV features without uncertainty modeling, limiting robustness to calibration errors and occlusions.

\vspace{-5mm}
\paragraph{Feedforward 4D Reconstruction.}
Feedforward 4D reconstruction directly regresses geometry and motion from image sequences in a single forward pass. 
DUSt3R and VGGT ~\cite{vggt,dust3r} recover dense 3D structure through transformer-based multi-view attention, while StreetForward~\cite{streetforward} and DGGT~\cite{dggt} extend this paradigm to dynamic scene understanding. STORM~\cite{storm} and EVolSplat4D~\cite{evolsplat4d} further explore dynamic volumetric and Gaussian-based 4D representations. Nevertheless, these methods remain deterministic and largely restricted to single-agent settings without uncertainty propagation.
\vspace{-5mm}
\paragraph{3D Gaussians for Driving and Collaboration.}
3D Gaussian Splatting~\cite{3dgs} provides a compact and differentiable scene representation using continuous Gaussian primitives. 
Recent works extend Gaussian representations to 3D detection~\cite{3dgs-det,gaussiandet}, occupancy prediction~\cite{huang2024gaussianformer,huang2024gsformer2}, and dynamic driving scenes through Street Gaussians and DrivingGaussian. VOGS-CP~\cite{chen2026vision} further explores Gaussian primitives for cooperative perception through learned cross-agent fusion. 
However, existing methods mainly focus on representation fidelity and rely on heuristic aggregation without explicit uncertainty modeling. 
In contrast, our framework extends Gaussian scene representations to cooperative 4D perception with uncertainty-aware Gaussian propagation in BEV space.
\vspace{-5mm}
\paragraph{Cooperative Multi-Object Tracking.}
Cooperative multi-object tracking has mainly been explored in LiDAR-based settings.
DMSTrack~\cite{chiu2024probabilistic} integrates differentiable Kalman filtering with learned observation covariance, while MOT-CUP\cite{su2024collaborative} employs conformal prediction for calibrated uncertainty propagation. 
However, these methods rely on discrete bounding-box representations and simplified uncertainty assumptions, limiting applicability to camera-only cooperative perception and failing to model anisotropic depth uncertainty from perspective projection.
\vspace{-20pt}

\section{Preliminaries}
\label{sec:dc}

Consider $N$ CAVs $\mathcal{S}$ within V2V range, each observing $F$ consecutive frames 
from $C$ cameras, $\mathbf{O}_i=\{\mathbf{I}^{i,c}_f \in \mathbb{R}^{H \times W \times 3}\}$,
with the goal of estimating a cooperative 4D world model, defined as follows:
\begin{equation}
    \mathcal{B}_i=\{(\boldsymbol{\mu}_k,\mathbf{v}_k,\theta_k,\mathbf{d}_k)\},
    \label{eq:world_model}
\end{equation}
where each tuple encodes position, velocity, heading, and dimensions of dynamic agents.
Existing methods either share dense features ($>$1~MB/frame) or exchange bounding boxes,
discarding uncertainty. CoCam4D addresses this via three subproblems.

\textcolor{navyblue}{\textbf{(P1) Per-Vehicle 4D Reconstruction.}}
Each vehicle independently reconstructs 3D Gaussians, velocities, and depth uncertainty
in a single forward pass:
\begin{equation}
    \left(\mathcal{G}_i, \mathbf{V}_i, \boldsymbol{\Sigma}_i\right) = f_{\mathrm{enc}}(\mathbf{O}_i).
    \label{eq:p1}
\end{equation}

\textcolor{navyblue}{\textbf{(P2) Compact Cooperative Messaging.}}
Each vehicle compresses its reconstruction into Dynamic Object Primitives (DOPs),
\begin{equation}
    \mathcal{M}_i = \left\{\mathcal{P}_k = \left(\boldsymbol{\mu}_k, \mathbf{v}_k,
    \theta_k, \mathbf{d}_k, \mathbf{R}_k\right)\right\}, \quad |\mathcal{M}_i| \leq \beta,
    \label{eq:p2}
\end{equation}
where $\mathbf{R}_k$ is the observation covariance derived analytically from $\boldsymbol{\Sigma}_i$
and camera geometry-encoding full geometric state and confidence in ${\approx}35$~bytes per object.

\textcolor{navyblue}{\textbf{(P3) Uncertainty-Weighted Cooperative Fusion.}}
Received DOPs are fused via an information filter,
\begin{equation}
    \mathbf{R}^{-1}_\mathrm{fused} = \mathbf{R}^{-1}_i + \mathbf{R}^{-1}_j,
    \label{eq:p3}
\end{equation}
which weights each observation by confidence without learned calibration. Anisotropic
covariances $\mathbf{R}_k$ ensure lateral views compensate for head-on depth uncertainty,
yielding detections $\mathcal{B}_i$ that exceed single-agent estimates.
\vspace{-15pt}

\section{Proposed Method}
\label{sec:method}


CoCam4D considers a set of $N$ connected autonomous vehicles (CAVs) $\mathcal{S}$ within V2V range, each equipped with $C$ surround-view cameras observing $F$ consecutive frames, and outputs at each vehicle a cooperative 4D world model $\mathcal{B}_i = \{(\boldsymbol{\mu}_k, \mathbf{v}_k, \theta_k, \mathbf{d}_k)\}$ with accurate cooperative 4D world model than any single vehicle can achieve alone. 
The pipeline consists of three stages, which visually described in Figure~\ref{fig:overview}: (i) per-vehicle feedforward 4D reconstruction with velocity-aware geometry and calibrated uncertainty; (ii) Dynamic Object Primitive (DOP) extraction that compresses representations into bandwidth-limited messages while preserving uncertainty structure; and (iii) cooperative fusion via an information filter that weights each vehicle according to complementary observation geometry. 
We described each stage, with full derivations, training details, and ablations provided in the supplementary material.
\vspace{-8pt}
\subsection{Per-Vehicle Feedforward 4D Reconstruction}
\label{subsec:per_vehicle}

Each vehicle processes its observations independently in a single forward pass, without iterative optimization or additional calibration refinement. 
The encoder $f_\mathrm{enc}$ produces a rich intermediate representation that feeds six specialized prediction heads, each addressing a distinct aspect of dynamic scene understanding.
\vspace{-10pt}
\paragraph{Tokenization and multi-scale attention.}

The $N_\mathrm{img}=F\cdot C$ input images are encoded by a frozen ViT backbone into patch tokens $\mathbf{X}\in\mathbb{R}^{B\times N_\mathrm{img}\times P\times D}$, where $P=(H/S)^2$, $P$ is number of patches (tokens) per image, $H$ is image height, and $S$ = patch size / stride of the ViT encoder. 
Standard global attention considers all frames symmetrically, making motion patterns less clear.
We instead use $L$ layers of alternating frame-wise attention (within each image over $P$ tokens) and global attention over all $N_\mathrm{img}\cdot P$ tokens following ~\eqref{eq:global_attn},
\begin{equation}
\mathbf{Z}=\mathrm{SelfAttn}\!\left(\mathrm{flatten}_{(n,p)}(\mathbf{X})\right), \quad 
\mathbf{Z}\in\mathbb{R}^{B\times (N_\mathrm{img}\cdot P)\times D}
\label{eq:global_attn}
\end{equation}

Deep layers capture scene-level geometry but lose fine detail, so we fuse with raw features via
\begin{equation}
\mathbf{F}_\mathrm{fused}=\mathrm{concat}(\mathbf{F}_\mathrm{raw},\mathbf{F}_\mathrm{attn}) \in \mathbb{R}^{B\times N_\mathrm{img}\times P\times 2D}
\label{eq:fused_feat}
\end{equation}

The fused feature $\mathbf{F}_\mathrm{fused}$ is used as the shared backbone for all six heads below.

\paragraph{Camera pose estimation.}

We follow the same calibration protocol as CoCa3D~\cite{coca3d}, using the standard dataset-provided camera intrinsics and extrinsics without additional offline cross-vehicle calibration refinement. The camera head $\mathcal{H}_\mathrm{cam}$ predicts per-image intrinsics $\mathbf{K}_n \in \mathbb{R}^{3 \times 3}$ and extrinsics $(\mathbf{R}_n, \mathbf{t}_n)$ for each image $n \in \{1,\ldots,N_\mathrm{img}\}$, with the first camera of the first frame defining the world origin. 
This formulation supports deployment across different vehicle platforms without requiring additional offline cross-vehicle calibration refinement.
\vspace{-10pt}
\paragraph{Depth and uncertainty prediction.}

The depth head $\mathcal{H}_\mathrm{dep}$ produces per-pixel metric depth $D_n \in \mathbb{R}^{H \times W}$ via VGGT-based multi-scale decoder equipped with skip connections from fused feature representations. 
Existing feedforward reconstruction methods~\cite{charatan2024pixelsplat,fan2024instantsplat} only output depth and do not predict calibrated uncertainty.
The uncertainty head $\mathcal{H}_\mathrm{unc}$, which is the first novel contribution of this work, shares the decoder architecture of $\mathcal{H}_\mathrm{dep}$ but independently trained in the manner of self-supervised way, and predicts per-pixel log-depth uncertainty $\hat{\Sigma}_n \in \mathbb{R}^{H \times W}$ with a softplus activation\cite{liu2016noisy} to ensure positivity. 
Supervision uses the prediction error from the depth head as an internally defined target:
\begin{equation}
    \mathcal{L}_\mathrm{unc} = \bigl\|\hat{\Sigma}_n - \mathrm{sg}\!\left(|D_n - D^*_n|\right)\bigr\|_1
    \label{eq:unc_loss}
\end{equation}
where $\mathrm{sg}(\cdot)$ denotes stop-gradient and $D^*_n$ is the ground-truth depth. 
This formulation does not require extra annotations beyond those used by the depth head, which makes uncertainty estimation a zero-cost addition during training.
\vspace{-10pt}
\paragraph{Velocity-aware Gaussian primitives.}

The Gaussian head $\mathcal{H}_\mathrm{gs}$ produces a set of 3D Gaussian primitives $\mathcal{G}_i = \{\mathbf{G}_k \in \mathbb{R}^d \mid k = 1,\ldots,P_\mathrm{gs}\}$. Each primitive encodes a local scene region through its mean $\boldsymbol{\mu}_k \in \mathbb{R}^3$, scale $\mathbf{s}_k \in \mathbb{R}^3$, rotation quaternion $\mathbf{r}_k \in \mathbb{R}^4$, opacity $a_k \in [0,1]$, color $\mathbf{c}_k \in \mathbb{R}^3$, and lifespan $\sigma_k \in \mathbb{R}^+$. The 3D center is back-projected from the head of depth output as:
\begin{equation}
    \boldsymbol{\mu}_n(\mathbf{u}) = \mathbf{R}_n^\top\!\left(\mathbf{K}_n^{-1}\,\tilde{\mathbf{u}}\,D_n(\mathbf{u}) - \mathbf{t}_n\right)
    \label{eq:backproject}
\end{equation}
where $\tilde{\mathbf{u}} = (u_x, u_y, 1)^\top$ is the homogeneous pixel coordinate. The covariance is constructed as $\boldsymbol{\Sigma}_k = \mathbf{R}_g\,\mathbf{S}\mathbf{S}^\top\mathbf{R}_g^\top$, with $\mathbf{S} = \mathrm{diag}(\mathbf{s}_k)$ and $\mathbf{R}_g = \mathrm{q2r}(\mathbf{r}_k)$. 

\paragraph{Connecting uncertainty to geometry.}

Crucially, the depth-axis scale is modulated by the uncertainty head output from ~\eqref{eq:unc_loss}:
\begin{equation}
    s_z = s_z^{\mathrm{base}} \times \left(1 + \beta\,\hat{\Sigma}_n(\mathbf{u})\right)
    \label{eq:scale_mod}
\end{equation}
amplifying the Gaussian extent along the viewing ray for uncertain observations. This modulation establishes the uncertainty chain end-to-end: depth uncertainty from ~\eqref{eq:unc_loss} propagates through ~\eqref{eq:scale_mod} to covariance, and ultimately to information filter fusion weighting in Section~\ref{subsec:coop}. 
Without this connection, cooperating vehicles consider confident side observations the same as uncertain front-facing views, which is a key limitation of methods that share object detections.

The lifespan parameter further modulates opacity over time:
\begin{equation}
    o_k(\mathbf{u}, t') = a_k(\mathbf{u})\exp\!\left(-\frac{1}{2}\,\frac{(t'-t)^2}{\sigma_k(\mathbf{u})}\right)
    \label{eq:lifespan}
\end{equation}
Static Gaussians can fade gracefully as their appearance becomes less reliable due to lighting or reflectance changes, as shown by the decay in ~\eqref{eq:lifespan}.
\vspace{-10pt}
\paragraph{Directed motion representation.}

Standard global attention considers all temporal frames symmetrically, which reduces the model ability to represent directed motion. 
The motion head $\mathcal{H}_\mathrm{motion}$ addresses this limitation by augmenting tokens with periodic time based temporal embeddings and applying a frame structured causal mask:
\begin{equation}
    \mathbf{M}[i,j] = \begin{cases}
    1 & \text{if } \mathrm{frame}(i) = f_s \text{ and } \mathrm{frame}(j) = f_s \pm 1, \\
    0 & \text{otherwise,}
    \end{cases}
    \label{eq:causal_mask}
\end{equation}
which restricts each query token at source frame $f_s$ to attend only to key tokens at adjacent frames. The masked attention in ~\eqref{eq:causal_mask} produces motion-aware features from which a multi-scale decoder regresses per-pixel forward and backward velocities $\mathbf{v}^+, \mathbf{v}^- \in \mathbb{R}^3$ and dynamic probability $p_\mathrm{dyn} \in [0,1]$. The velocity covariance is derived analytically from the uncertainty head, propagating the depth-scaling law through to motion:
\begin{equation}
    \sigma^2_v \propto \frac{\sigma^2_\mathrm{depth}}{\Delta t^2} \cdot \frac{D^2}{f^2}
    \label{eq:vel_cov}
\end{equation}
This covariance from ~\eqref{eq:vel_cov} later drives the information filter weighting in ~\eqref{eq:if_state}.

\textbf{Background modeling.}
The sky head $\mathcal{H}_\mathrm{sky}$ represents unbounded background regions with a fixed set of Gaussians whose centers are sampled on a upper half-sphere of radius $r_\mathrm{sky}$.
Their colors are initialized by projecting image pixels onto the hemisphere and refined by a lightweight MLP, preventing the depth head from wasting capacity on infinitely distant sky pixels.

\textbf{Dynamic-static decomposition and rendering.}

Pixels with $p_\mathrm{dyn} > \tau = 0.5$ contribute dynamic Gaussians $\mathcal{G}_\mathrm{dyn}$; the remainder contribute static Gaussians $\mathcal{G}_\mathrm{static}$ accumulated across all frames. The full Gaussian set used to render frame $f$ is:
\begin{equation}
    \hat{\mathcal{G}}^f = \mathcal{G}_\mathrm{static} \cup \bigcup_{t \neq f} \mathcal{G}^{f \leftarrow t}_\mathrm{dyn} \cup\, \mathcal{G}_\mathrm{sky}
    \label{eq:render_set}
\end{equation}
where $\mathcal{G}^{f \leftarrow t}_\mathrm{dyn}$ denotes dynamic Gaussians from frame $t$ warped into frame $f$ using predicted velocities. By excluding dynamic Gaussians at frame $f$ itself, as shown in ~\eqref{eq:render_set}, the model is forced to explain dynamic content through adjacent-frame velocity predictions, enforcing motion correctness via photometric loss without any explicit flow supervision—a form of self-consistency regularization absent from static reconstruction methods. The rendered image is:
\begin{equation}
    \hat{\mathbf{I}}^f = \mathrm{Renderer}\!\left(\hat{\mathcal{G}}^f,\, \boldsymbol{\Pi}^f\right)
    \label{eq:render}
\end{equation}

\textbf{Training objective.}

The overall per-vehicle training objective combines photometric, opacity, dynamic mask, lifespan, and uncertainty terms:
\begin{equation}
    \mathcal{L}_\mathrm{vehicle} = \mathcal{L}_\mathrm{rgb} + \lambda_\alpha \mathcal{L}_\mathrm{opacity} + \lambda_d \mathcal{L}_\mathrm{dynamic} + \lambda_\sigma \mathcal{L}_\mathrm{lifespan} + \lambda_u \mathcal{L}_\mathrm{unc}
    \label{eq:total_loss}
\end{equation}

\subsection{Dynamic Object Primitive Extraction}
\label{subsec:dop}

Each vehicle generates compact object-level representations from its output. Existing feature sharing methods\cite{chen2019fcooper, xu2022opv2v, v2xvit, cobevt}transmit raw or intermediate features, but they require very high bandwidth (more than 1 MB per frame). In contrast, detection-sharing methods exchange only the final bounding boxes, which reduces communication cost but loses important uncertainty information required for reliable fusion. DOPs provide a balanced solution by using only about 35 bytes per primitive while still preserving the full observation covariance matrix $\mathbf{R}_k$, which captures geometric confidence.

\textbf{Clustering and filtering.}
Connected-component analysis is applied to the binary dynamic mask ($p_\mathrm{dyn} > \tau$) to group pixels into clusters ${\mathcal{C}k}$. Small components with fewer than $N\mathrm{min} = 20$ pixels are removed as noise. In addition, components with a 3D extent larger than 15m are split because they are likely to contain multiple objects. This simple heuristic helps prevent different objects, such as buses and motorcycles, from being merged into a single primitive.

\textbf{Primitive construction.}
For each surviving cluster $\mathcal{C}_k$, we form a Dynamic Object Primitive (DOP):
\begin{equation}
    \mathcal{P}_k = \left(\boldsymbol{\mu}_k,\, \mathbf{v}_k,\, \theta_k,\, \mathbf{d}_k,\, \mathbf{R}_k\right),
    \label{eq:dop}
\end{equation}
where $\boldsymbol{\mu}_k$ is the 3D centroid back-projected via ~\eqref{eq:backproject}, $\mathbf{v}_k$ is the mean cluster velocity, $\theta_k$ is the heading derived from $\mathbf{v}_k$, $\mathbf{d}_k$ encodes object dimensions from eigen decomposition of the cluster covariance, and $\mathbf{R}_k$ is the observation covariance.

\textbf{Analytical uncertainty propagation.}
%
The observation covariance in ~\eqref{eq:dop} follows the geometry of the camera system. As the object distance increases, the depth uncertainty grows quadratically, while the lateral uncertainty increases linearly. As a result, the uncertainty forms an anisotropic ellipsoid.

\begin{equation}
    \mathbf{R}_k = \mathrm{diag}\!\left(\sigma^2_x(z),\, \sigma^2_y(z),\, \sigma^2_z(z)\right), \quad \sigma^2_z(z) \propto \frac{z^4}{f^2},
    \label{eq:obs_cov}
\end{equation}
where $f$ is the focal length and $z$ is the distance of object from the camera.
This quadratic scaling is not learned from data. Instead, it is analytically derived from the pinhole camera model together with the depth uncertainty $\hat{\Sigma}_n$ predicted by the uncertainty head.
When a vehicle observes an object from the side at a distance of $10$m, the depth uncertainty is low ($\sigma_z \approx 0.5$,m). However, when the same object is observed from the front at the same distance, the depth uncertainty becomes much higher ($\sigma_z \approx 2.0$m).
This complementary structure is the foundation of cooperative gain.

\textbf{Quantization and bandwidth compliance.}
Each DOP from ~\eqref{eq:dop} is quantized to $\approx$35\,bytes: 12 bytes for $\boldsymbol{\mu}_k$ (3×4-byte floats), 6 bytes for $\mathbf{v}_k$ (3×2-byte ints), 2 bytes for $\theta_k$, 6 bytes for $\mathbf{d}_k$, and 9 bytes for the diagonal $\mathbf{R}_k$. The total message budget satisfies:
\begin{equation}
    |\mathcal{M}_i| \leq \beta \quad \forall\, i \in \mathcal{S},
    \label{eq:budget}
\end{equation}
where $\beta$ is the C-V2X sidelink capacity at 10\,Hz ($\approx$1400\,bytes), allowing $\sim$40 objects per vehicle per frame.
\vspace{-15pt}

\subsection{Cooperative Fusion}
\label{subsec:coop}

Upon receiving messages ${\mathcal{M}j}{j \neq i}$ from other cooperating vehicles, it first aligns and matches the DOPs, and then fuses the matched ones to produce detections $\mathcal{B}_i$. These fused results are more accurate than those from any single vehicle alone. 
The fusion process is purely geometric, with no learned calibration network, no multi-agent training, and no gradient flow between vehicles.

\textbf{Spatial alignment.}
Each vehicle knows the relative transform $T_{ij}(\mathbf{x}) = \mathbf{U}_{ij}\mathbf{x} + \mathbf{t}_{ij}$ from cooperator $j$ to ego $i$ via GPS/IMU. For a DOP with centroid $\boldsymbol{\mu}$ and covariance $\mathbf{R}$ in the $j$th frame, the transformed parameters in the $i$th frame are:
\begin{align}
    \boldsymbol{\mu}' &= \mathbf{U}_{ij}\,\boldsymbol{\mu} + \mathbf{t}_{ij}, \label{eq:mean_xf} \\
    \mathbf{R}' &= \mathbf{U}_{ij}\,\mathbf{R}\,\mathbf{U}_{ij}^\top, \label{eq:cov_xf}
\end{align}
%
This transformation rotates the uncertainty ellipsoid without changing its axis lengths, which makes cross-vehicle alignment simple, closed-form, and easy to interpret. After alignment, only the DOPs whose transformed centroids lie inside the ego vehicle’s region of interest $\mathbf{ROI}_i$ are kept. This significantly reduces the computational cost.

\textbf{Latency compensation.}
Communication delays $\Delta t_\mathrm{comm}$ mean that the received DOPs correspond to object positions from the past. Instead of discarding these outdated observations, we update each received DOP by forward-predicting its current position using the embedded velocity in ~\eqref{eq:latency}.
\begin{equation}
    \boldsymbol{\mu}_\mathrm{comp} = \boldsymbol{\mu} + \mathbf{v}\,\Delta t_\mathrm{comm},
    \label{eq:latency}
\end{equation}
%
At the same time, the covariance $\mathbf{R}k$ is increased using a constant-velocity process noise term $\mathbf{Q},\Delta t_{comm}$ to reflect the higher uncertainty in position. This compensation in ~\eqref{eq:latency} is done analytically and does not rely on any learned warping network, which helps the method remain robust under varying network delays.
\textbf{Cross-vehicle data association.}
Ego DOPs are matched against cooperator DOPs via the Hungarian algorithm applied to a Mahalanobis cost matrix which is as follows:
\begin{equation}
    d^2(\mathcal{P}_A, \mathcal{P}_B) = (\boldsymbol{\mu}_A - \boldsymbol{\mu}_B)^\top (\mathbf{R}_A + \mathbf{R}_B)^{-1} (\boldsymbol{\mu}_A - \boldsymbol{\mu}_B),
    \label{eq:mahal}
\end{equation}
%
where the position difference is weighted by the combined observation covariance of the two DOPs. Pairs exceeding a chi-squared gate $\chi^2_{3,0.95}=7.815$ are rejected as implausible associations. 
Including the velocity term helps avoid incorrect matching between static and dynamic vehicles that are close to each other. This is a common failure in methods that use only position, and it becomes especially important in dense urban scenes with many stationary vehicles.

\textbf{Information filter fusion.}
For each matched DOP pair, we perform cooperative fusion via the information filter's additive update. Define the information matrix $\mathbf{Y}_i = \mathbf{R}_i^{-1}$ and information state $\hat{\mathbf{y}}_i = \mathbf{R}_i^{-1}\boldsymbol{\mu}_i$. For $|\mathcal{S}|$ cooperating vehicles, the fused estimate is:
\begin{align}
    \mathbf{Y}_\mathrm{fused} &= \mathbf{Y}_\mathrm{prior} +
    \sum_{j \in \mathcal{S}} \mathbf{H}_j^\top \mathbf{R}_j^{-1}
    \mathbf{H}_j, \label{eq:if_update} \\
    \hat{\mathbf{y}}_\mathrm{fused} &= \hat{\mathbf{y}}_\mathrm{prior} +
    \sum_{j \in \mathcal{S}} \mathbf{H}_j^\top \mathbf{R}_j^{-1}
    \mathbf{z}_j, \label{eq:if_state}
\end{align}
where $\mathbf{H}_j$ is the observation matrix and $\mathbf{z}_j$ is the measurement from vehicle $j$. The updates in \eqref{eq:if_state} are purely additive and order-independent, naturally supporting decentralized operation with any number of vehicles where no fusion center, no coordinator, no global consensus protocol. The fused position and covariance are recovered as:
\begin{equation}
    \boldsymbol{\mu}_\mathrm{fused} = \mathbf{Y}_\mathrm{fused}^{-1}\,\hat{\mathbf{y}}_\mathrm{fused}, \qquad \mathbf{R}_\mathrm{fused} = \mathbf{Y}_\mathrm{fused}^{-1}.
    \label{eq:if_recover}
\end{equation}

\textbf{Bayes-optimality of the update.}
The additive rule in \eqref{eq:if_update}--\ref{eq:if_recover} is not a heuristic:
for independent Gaussian observations $\boldsymbol{\mu}_j\!\sim\!\mathcal{N}(\mathbf{x},\mathbf{R}_j)$
of a common state $\mathbf{x}$, minimising the negative log-likelihood
$\tfrac{1}{2}\sum_j(\boldsymbol{\mu}_j-\mathbf{x})^\top\mathbf{R}_j^{-1}(\boldsymbol{\mu}_j-\mathbf{x})$
yields the maximum-likelihood estimate
$\hat{\mathbf{x}}=(\sum_j\mathbf{R}_j^{-1})^{-1}(\sum_j\mathbf{R}_j^{-1}\boldsymbol{\mu}_j)$
with covariance $(\sum_j\mathbf{R}_j^{-1})^{-1}$, which is exactly
\eqref{eq:if_recover}. Thus CoCam4D's fusion is the Bayes-optimal estimator under
its noise model, with covariances derived analytically from perspective geometry
rather than learned; the full derivation is given in the supplementary.

\textbf{Automatic complementary weighting.}
For two vehicles observing the same object, Vehicle~A sees it from the side at a $5$m lateral offset with low depth uncertainty ($\sigma_z = 0.3$ m), while Vehicle ~B observes it head-on at $15$m with much higher uncertainty ($\sigma_z = 3.5$ m). Since information weight is inversely proportional to the square of uncertainty, Vehicle ~A provides a much higher weight ($\mathbf{Y}_A \approx 11.1$) compared to Vehicle B ($\mathbf{Y}_B \approx 0.08$). As a result, Vehicle A contributes about $\sim$140$\times$ more depth information due to its better viewing geometry.
This explains the gain of cooperative perception over single-agent methods.

\textbf{Shape completion via Gaussian union.}
Beyond fusing object states, we merge the raw Gaussian shape representations of matched objects. For each matched pair, we take the union of dynamic Gaussians from both vehicles transformed via ~\eqref{eq:mean_xf} and \eqref{eq:cov_xf} to the fused centroid from ~\eqref{eq:if_recover}, prioritizing Gaussians from the vehicle with lower aggregate uncertainty $\mathrm{tr}(\mathbf{R}_k)$, which is defined as follows:
\begin{equation}
    \mathcal{G}^{\mathrm{obj}}_\mathrm{fused} = \mathcal{G}^A_\mathrm{dyn} \cup \mathcal{G}^B_\mathrm{dyn},
    \label{eq:shape_union}
\end{equation}
producing geometrically more complete object representations. 
When Vehicle~A observes only the rear of a detected vehicle and Vehicle~B observes only its front, their combined information in ~\eqref{eq:shape_union} reconstructs the full object geometry. This level of shape completion is not possible for prior cooperative perception methods that rely only on bounding boxes.

\vspace{-20pt}

\section{Experiments}
\vspace{-2mm}
\begin{figure}[t] 
    \centering
    \includegraphics[width=.85\textwidth]{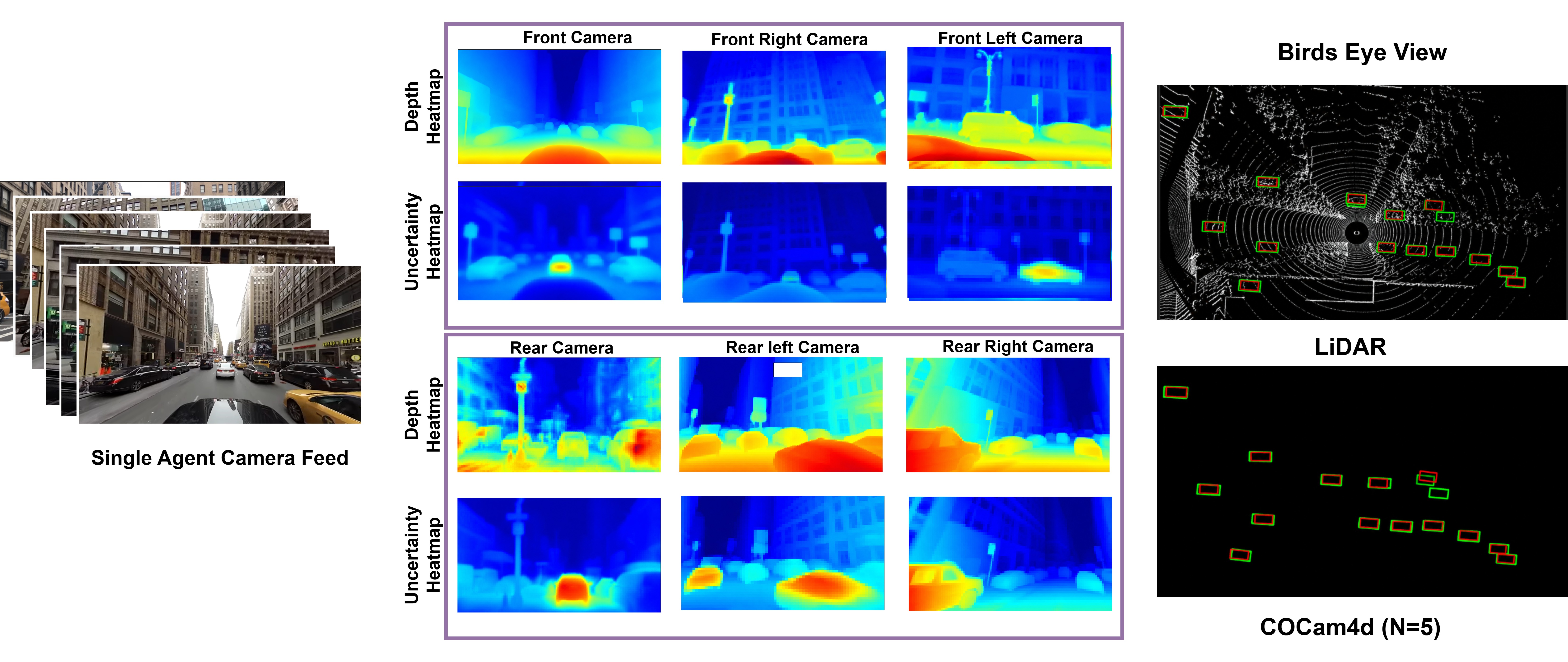}
    \vspace{-4mm}
    \caption{\textbf{Single-agent perception outputs in CoCam4D.} The figure shows surround-view inputs from six vehicle-mounted cameras along with their corresponding predicted depth and uncertainty heatmaps. The right column compares LiDAR observations with CoCam4D cooperative object predictions (N=5)}
    \label{fig:vis_results}
    \vspace{-6mm}
\end{figure}



We evaluate CoCam4D on two datasets that cover simulated and real-world cooperative driving scenarios, namely OPV2V+~\cite{hu2023collaboration} and DAIR-V2X-C~\cite{yu2022dair}. Table \ref{tab:configs} shows the configurations of the experiments.
\vspace{-10pt}



\begin{table}[h]
    \centering
    \footnotesize
    \setlength{\tabcolsep}{3pt}
    \resizebox{0.6\linewidth}{!}{
    \begin{tabular}{clccc}
        \toprule
        \textbf{ID} & \textbf{Name} & \textbf{Sensor} &
        \textbf{Vehicles} & \textbf{Purpose} \\
        \midrule
        C1 & Single-Vehicle Camera & 4$\times$ RGB & 1 & Lower bound \\
        C2 & Single-Vehicle LiDAR & 64-beam & 1 & Target to match \\
        C3 & CoCam4D ($N$=2) & 4$\times$ RGB & 2 & Min. cooperation \\
        C4 & CoCam4D ($N$=5) & 4$\times$ RGB & 5 & Main result \\
        C5 & Cooperative LiDAR & 64-beam & 2 & Upper bound \\
        \bottomrule
    \end{tabular}
    }
    \caption{Experimental configurations used in the evaluation.}
    \label{tab:configs}
\end{table}

\subsection{Evaluation Metrics}

For 3D detection we report AP@0.3/0.5/0.7 in both BEV and 3D, broken
down per range bin (0--30\,m, 30--50\,m, 50--70\,m, 70--100\,m) and per
class (Car, Pedestrian, Cyclist). For scene reconstruction we report
Chamfer-L1 and F-score@0.5\,m against LiDAR ground truth on real-world
data. Novel view synthesis quality is measured by PSNR and SSIM following
the protocol of StreetForward~\cite{streetforward}. Dynamic tracking
performance is evaluated with AMOTA and AMOTP on the OPV2V+\cite{hu2023collaboration} tracking
split. Communication cost is measured in KB/s per vehicle pair at 10\,Hz,
and end-to-end latency is reported in milliseconds from image capture to
fused detections.
\vspace{-15pt}
\subsection{Experimental Results}





We compare CoCam4D with representative cooperative and single-vehicle baselines across all configurations. \textbf{Cooperative camera baselines} include CoCa3D~\cite{coca3d}, our primary and closest prior work. IFTR~\cite{wang2024iftr}, which performs object-level instance query fusion whereas Where2comm~\cite{hu2022where2comm}, which uses confidence-aware spatial feature selection. V2X-ViT~\cite{xu2022v2xvit}, based on transformer-driven cross-agent fusion and VOGS-CP~\cite{chen2026vision}, which transmits 3D Gaussian primitives for cooperative semantic occupancy. 
\textbf{Single-vehicle baselines} include BEVFormer~\cite{bevformer} for camera-only perception (C1) and LiDAR-based CenterPoint as the single-vehicle LiDAR reference (C2). All baselines use official implementations and reported hyperparameters without re-tuning. Figure ~\ref{fig:vis_results} shows the single-agent perception results of the proposed CoCam4D, including the depth map, uncertainty heatmap, and detected 3D bounding boxes.
\vspace{-15pt}

\paragraph{Primary Results.}

\begin{table*}[t]
    \centering
    \scriptsize
    \setlength{\tabcolsep}{3pt}
    \renewcommand{\arraystretch}{1.05}

    \resizebox{0.6\textwidth}{!}{
    \begin{tabular}{lcccccccc}
        \toprule
        \multirow{2}{*}{\textbf{Method}} &
        \multirow{2}{*}{\textbf{Modality}} &
        \multirow{2}{*}{\textbf{Coop.}} &
        \multirow{2}{*}{$N$} &
        \multicolumn{2}{c}{\textbf{AP@0.5 BEV}} &
        \multirow{2}{*}{\textbf{NDS}} &
        \multirow{2}{*}{\textbf{AMOTA}} &
        \multirow{2}{*}{\textbf{KB/s}} \\
        \cmidrule(lr){5-6}
        & & & & 0--50\,m & 50--100\,m & & & \\
        \midrule

        CenterPoint & LiDAR & \xmark & 1 &
        62.3 & 35.8 & 52.1 & 45.7 & --- \\

        BEVFormer~\cite{bevformer} & Camera & \xmark & 1 &
        38.4 & 14.7 & 33.2 & 24.6 & --- \\

        \midrule

        Where2comm~\cite{hu2022where2comm} & Camera & \cmark & 2 &
        44.2 & 18.9 & 37.8 & 29.4 & 840 \\

        V2X-ViT~\cite{xu2022v2xvit} & Camera & \cmark & 2 &
        46.8 & 21.3 & 40.1 & 31.7 & 920 \\

        CoCa3D~\cite{coca3d} & Camera & \cmark & 2 &
        51.4 & 25.6 & 44.3 & 36.2 & 1240 \\

        VOGS-CP~\cite{chen2026vision} & Camera & \cmark & 2 &
        49.7 & 23.8 & 42.6 & 34.5 & 386 \\

        \midrule

        CoCam4D ($N$=2) & Camera & \cmark & 2 &
        57.3 & 28.5 & 48.7 & 40.2 & 58 \\

        CoCam4D ($N$=5) & Camera & \cmark & 5 &
        \textcolor{cyan}{\textbf{65.8}} &
        \textcolor{cyan}{\textbf{38.6}} &
        \textcolor{cyan}{\textbf{54.3}} &
        \textcolor{cyan}{\textbf{48.1}} & 58 \\

        \midrule

        Cooperative LiDAR & LiDAR & \cmark & 2 &
        71.4 & 44.2 & 60.8 & 56.3 & --- \\

        \bottomrule
    \end{tabular}
    }

    \caption{Simulation results on on OPV2V+\cite{hu2023collaboration}, the extended version of the original dataset OPV2V. \textbf{Bold} entries exceed
    single-vehicle LiDAR (CenterPoint). CoCam4D ($N$=5) is the first
    cooperative camera method to surpass single-vehicle LiDAR on all
    reported metrics. CoCam4D ($N$=2) outperforms CoCa3D by 11.5\%
    AP@0.5 at 0--50\,m and 11.3\% at 50--100\,m while transmitting
    21$\times$ less bandwidth.}
    \label{tab:main}
    \vspace{-3mm}
\end{table*}

Table~\ref{tab:main} summarizes results on OPV2V+ \cite{hu2023collaboration}. 
Single-vehicle camera (BEVFormer) trails single-vehicle LiDAR (CenterPoint) by 23.9 points at 0--50,m and 21.1 points at 50--100,m, reflecting the quadratic growth of depth uncertainty with range in ~\eqref{eq:obs_cov}. 
Cooperative camera methods reduce this gap through shared BEV features, but without uncertainty modeling they weight all observations equally, limiting long-range gains. VOGS-CP improves representation with Gaussian primitives but lacks velocity estimation and uncertainty-aware fusion, remaining below CoCa3D in AP. 
CoCam4D at $N$=2 surpasses CoCa3D by 11.5\% at close range and 11.3\% at long range, where the depth-dependent weighting in ~\eqref{eq:if_state} better exploits complementary viewpoints. 
At $N$=5, CoCam4D outperforms single-vehicle LiDAR across all metrics while requiring only 58\,KB/s per vehicle pair, remaining within C-V2X sidelink limits.
\vspace{-10pt}

\paragraph{Comparison on Real-World Data.}

\begin{table}[h]
\vspace{-4mm}
    \centering
    \footnotesize
    \resizebox{0.42\linewidth}{!}{
    \begin{tabular}{lcc}
        \toprule
        \multirow{2}{*}{\textbf{Method}} &
        \multicolumn{2}{c}{\textbf{DAIR-V2X-C\cite{yu2022dair}}} \\
        \cmidrule(lr){2-3}
        & AP@0.5$\uparrow$ &
        AP@0.7$\uparrow$ \\
        \midrule
        BEVFormer~\cite{bevformer} &
        28.7 & 18.4 \\
        
        Where2comm~\cite{hu2022where2comm} &
        34.2 & 21.8 \\
        
        CoCa3D~\cite{coca3d} &
        38.6 & 24.3 \\
        
        \textcolor{cyan}{CoCam4D (Ours)} &
        \textcolor{cyan}{\textbf{42.7}} &
        \textcolor{cyan}{\textbf{27.0}} \\
        \bottomrule
    \end{tabular}
    }
    \caption{Comparative results on the real-world dataset DAIR-V2X-C\cite{yu2022dair}. CoCam4D
    generalizes to real sensor data without domain-specific fine-tuning,
    improving AP@0.5 by 10.6\% over CoCa3D.}
    \label{tab:realworld}
\end{table}

Table~\ref{tab:realworld} validates generalization to real sensor data.
On DAIR-V2X-C\cite{yu2022dair}, CoCam4D improves AP@0.5 by 10.6\% over CoCa3D and
generalizes to vehicle-infrastructure cooperation despite training exclusively on V2V
data, indicating that the information filter in
~\eqref{eq:if_state} and \eqref{eq:if_recover} is agnostic to whether
the cooperating agent is a vehicle or a fixed infrastructure node, as long
as its observation covariance is well calibrated. The gains persist under real-world
calibration noise because the depth-dependent covariance $\mathbf{R}_k\!\propto\!z^4/f^2$
down-weights uncertain far-range observations analytically rather than through a
learned module, so cross-sensor domain shift does not require re-tuning a fusion network.
\vspace{-8pt}

\paragraph{Cooperative Density Sweep.}
\label{subsec:density}

\begin{table}[h]
\vspace{-6mm}
    \centering
    \footnotesize
    \resizebox{0.45\linewidth}{!}{
    \begin{tabular}{ccccc}
        \toprule
        $N$ & \textbf{AP@0.5$\uparrow$} &
        \textbf{AP@0.7$\uparrow$} &
        \textbf{AMOTA$\uparrow$} &
        \textbf{KB/s} \\
        \midrule
        1 (camera) & 35.2 & 18.4 & 26.8 & --- \\
        1 (LiDAR)  & 62.3 & 38.6 & 45.7 & --- \\
        \midrule
        2 (cameras) & 48.6 & 28.3 & 36.4 & 42 \\
        3 (cameras) & 54.8 & 33.6 & 41.2 & 58 \\
        5 (cameras) &
        \textcolor{cyan}{\underline{63.7}} &
        \textcolor{cyan}{\underline{40.2}} &
        \textcolor{cyan}{\underline{47.8}} & 87 \\
        10 (cameras) & 68.4 & 43.8 & 51.4 & 156 \\
        20 (cameras) & 71.2 & 45.6 & 53.8 & 289 \\
        \bottomrule
    \end{tabular}
    }
    \caption{Cooperative density sweep on CARLA. Underlined row marks
    the critical density threshold at which CoCam4D first exceeds
    single-vehicle LiDAR (shown in the dashed reference row) on all
    three metrics. Performance scales consistently with $N$ through
    diminishing returns above $N$=10.}
    \label{tab:density}
\end{table}

\begin{figure}[t] 
    \centering
    \includegraphics[width=\textwidth]{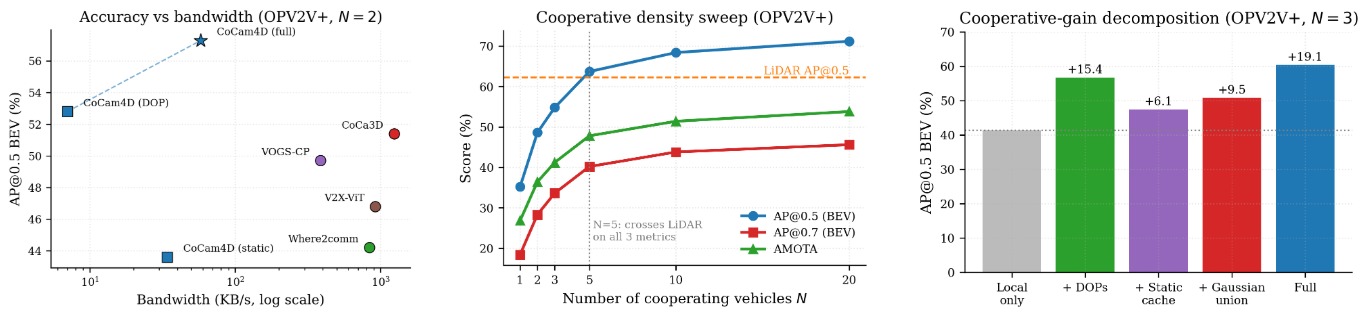}
    \vspace{-5mm}
    \caption{Cooperative perception analysis on OPV2V+. Left: accuracy-bandwidth trade-off. Middle: performance scaling with cooperative density. Right: contribution of DOPs, Gaussian caching, and uncertainty-aware fusion to cooperative gain.}
    \label{fig:graph}
    \vspace{-4mm}
\end{figure}

Table~\ref{tab:density} presents the first systematic study of the
cooperative density threshold. This sweep is conducted on CARLA, where the number
of cooperating agents can be controlled precisely; absolute values therefore differ
from the OPV2V+ results reported elsewhere in this section. CoCam4D at $N$=5 achieves 63.7\%
AP@0.5, 40.2\% AP@0.7, and 47.8 AMOTA, crossing the single-vehicle
LiDAR reference on all three metrics simultaneously and directly
answering the central question of this work: \emph{five cooperative camera vehicles achieve performance comparable to a single LiDAR-equipped vehicle}, discussed in figure~\ref{fig:graph}(b). The performance curve is nearly linear from
$N$=1 to $N$=5, reflecting the information filter's correct exploitation
of each new viewpoint through independent depth-uncertainty contributions.
Beyond $N$=10, diminishing returns set in as overlapping fields of view
provide increasingly redundant lateral coverage rather than complementary
depth triangulation, and bandwidth grows proportionally with $N$.
\vspace{-15pt}

\subsection{Ablation Studies}

All ablations are conducted on OPV2V+\cite{hu2023collaboration} with $N$=3 cooperative vehicles.

\begin{table}[h]
    \centering
    \footnotesize
    \resizebox{0.6\linewidth}{!}{
    \begin{tabular}{lccccccc}
        \toprule
        \textbf{Variant} &
        $\mathcal{H}_\mathrm{unc}$ &
        \textbf{Causal} &
        \textbf{Lifespan} &
        \textbf{IF} &
        \textbf{AP@0.5$\uparrow$} &
        \textbf{AMOTA$\uparrow$} &
        \textbf{KB/s} \\
        \midrule
        \textcolor{cyan}{Full CoCam4D} &
        \cmark & \cmark & \cmark & \cmark &
        \textcolor{cyan}{\textbf{60.4}} &
        \textcolor{cyan}{\textbf{44.8}} & 58 \\
        
        w/o $\mathcal{H}_\mathrm{unc}$ &
        \xmark & \cmark & \cmark & \cmark &
        55.2 & 40.3 & 58 \\
        
        w/o Causal Attn. &
        \cmark & \xmark & \cmark & \cmark &
        56.8 & 37.6 & 58 \\
        
        w/o Lifespan &
        \cmark & \cmark & \xmark & \cmark &
        58.1 & 43.2 & 58 \\
        
        BEV fusion &
        \cmark & \cmark & \cmark & \xmark &
        53.7 & 39.8 & 186 \\
        
        Late fusion &
        \cmark & \cmark & \cmark & \xmark &
        49.2 & 35.6 & 12 \\
        
        w/o $\mathbf{v}$ in assoc. &
        \cmark & \cmark & \cmark & \cmark &
        58.6 & 39.4 & 58 \\
        \bottomrule
    \end{tabular}
    }
    \caption{Ablation study on OPV2V+\cite{hu2023collaboration} ($N$=3). IF = information filter;
    BEV fusion replaces the Gaussian product with learned BEV attention;
    Late fusion transmits bounding boxes only.}
    \label{tab:ablation}
    \vspace{-4mm}
\end{table}

\textbf{Effect of the Uncertainty Head.} Removing $\mathcal{H}_\mathrm{unc}$ enforces uniform covariance in the information filter, eliminating depth-dependent weighting in ~\eqref{eq:if_update}. This causes the largest degradation (5.2 AP, 4.5 AMOTA), confirming that the uncertainty pipeline from ~\eqref{eq:unc_loss} through ~\eqref{eq:scale_mod} to fusion is the primary source of gains over prior CP methods.

\textbf{Effect of Causal Attention.} Replacing causal masked attention with symmetric global attention reduces AMOTA by 7.2 and AP by 3.6. The stronger tracking degradation confirms that directed attention in ~\eqref{eq:causal_mask} is essential for temporal velocity consistency, while detection is less affected since the backbone still preserves depth cues.

\textbf{Effect of Lifespan.} Removing lifespan $\sigma_k$ lowers AP by 2.3 due to stale static Gaussian accumulation. The smaller impact suggests lifespan mainly improves static scene quality, while DOP-driven dynamic detection remains largely unaffected.

\textbf{Gaussian Fusion vs.\ BEV Fusion.} Replacing the information filter with learned BEV attention fusion reduces AP@0.5 by 6.7 and increases bandwidth from 58 to 186\,KB/s, showing Gaussian product fusion is both more accurate and efficient. Late bounding-box fusion further reduces AP by 4.5, highlighting the importance of intermediate Gaussian structure.

\textbf{Velocity Term in Association.} Removing velocity from the Mahalanobis cost in ~\eqref{eq:mahal} slightly lowers AP (1.8) but significantly degrades AMOTA by 5.4, confirming that velocity-aware association is critical for identity consistency in dynamic scenes, particularly at intersections.
\vspace{-8pt}
\paragraph{Robustness Analysis}

\begin{table}[h]
    \centering
    \small
    \resizebox{0.5\linewidth}{!}{
    \begin{tabular}{lcccc}
        \toprule
        & \multicolumn{2}{c}{\textbf{Latency (ms)}} &
        \multicolumn{2}{c}{\textbf{Pose Noise (m STD)}} \\
        \cmidrule(lr){2-3}\cmidrule(lr){4-5}
        \textbf{Method} & 100 & 500 & 0.2 & 1.0 \\
        \midrule
        Where2comm~\cite{hu2022where2comm} &
        44.3 & 32.8 & 46.2 & 38.7 \\
        CoCa3D~\cite{coca3d} &
        49.8 & 38.4 & 50.6 & 43.2 \\
        \textcolor{cyan}{CoCam4D (Ours)} &
        \textcolor{cyan}{\textbf{56.4}} &
        \textcolor{cyan}{\textbf{53.7}} &
        \textcolor{cyan}{\textbf{58.9}} &
        \textcolor{cyan}{\textbf{55.4}} \\
        \bottomrule
    \end{tabular}
    }
    \caption{AP@0.5 BEV under communication latency and relative pose
    noise on OPV2V+\cite{hu2023collaboration} ($N$=3). CoCam4D degrades by only 2.7 AP under
    500\,ms latency versus 11.4 for CoCa3D, owing to velocity-based
    compensation in ~\eqref{eq:latency}.}
    \label{tab:robustness}
\end{table}

Table~\ref{tab:robustness} reveals two distinct robustness advantages.
Under 500\,ms communication latency, CoCam4D degrades by only 2.7 AP
because each received DOP carries an embedded velocity that enables
analytical forward prediction via ~\eqref{eq:latency}, whereas BEV
feature methods that transmit static snapshots degrade by 11.4 AP
(CoCa3D) and 11.5 AP (Where2comm) as their transmitted features become
stale. Under 1.0\,m pose noise STD, CoCam4D retains 55.4 AP while
CoCa3D drops to 43.2. The covariance inflation mechanism absorbs the
additional localization uncertainty directly into $\mathbf{R}_k$,
which the information filter in ~\eqref{eq:if_state} then correctly
down-weights, rather than propagating corrupted geometric features into
a learned fusion module that has no mechanism to distinguish reliable
from unreliable observations.
\vspace{-15pt}

\subsection{Communication Efficiency}

\begin{table}[h]
\vspace{-4mm}
    \centering
    \small
    \resizebox{0.6\linewidth}{!}{
    \begin{tabular}{llccc}
        \toprule
        \textbf{Method} & \textbf{Repr.} &
        \textbf{KB/s$\downarrow$} &
        \textbf{AP@0.5$\uparrow$} &
        \textbf{AP/KB$\uparrow$} \\
        \midrule
        Where2comm~\cite{hu2022where2comm} &
        BEV & 840 & 44.2 & 0.053 \\
        V2X-ViT~\cite{xu2022v2xvit} &
        BEV & 920 & 46.8 & 0.051 \\
        CoCa3D~\cite{coca3d} &
        BEV & 1240 & 51.4 & 0.041 \\
        VOGS-CP~\cite{chen2026vision} &
        Gaussians & 386 & 49.7 & 0.129 \\
        \midrule
        \textcolor{cyan}{CoCam4D (static)} &
        \textcolor{cyan}{Gaussians} & \textcolor{cyan}{34} & \textcolor{cyan}{43.6} & \textcolor{cyan}{1.282} \\
        \textcolor{cyan}{CoCam4D (DOP)} &
        \textcolor{cyan}{DOPs} & \textcolor{cyan}{7} & \textcolor{cyan}{52.8} & \textcolor{cyan}{7.543} \\
        \textcolor{cyan}{\textbf{CoCam4D (full)}} &
        \textcolor{cyan}{G$+$DOPs} &
        \textcolor{cyan}{\textbf{58}} &
        \textcolor{cyan}{\textbf{57.3}} &
        \textcolor{cyan}{\textbf{0.988}} \\
        \bottomrule
    \end{tabular}
    }
    \caption{Communication efficiency at $N$=2. AP/KB measures
    detection accuracy per unit bandwidth. CoCam4D (full) achieves
    the highest AP while transmitting 21$\times$ less than CoCa3D.
    Static Gaussians are transmitted once and amortized; the 58\,KB/s
    figure reflects steady-state per-frame DOP cost plus the amortized
    static cache.}
    \label{tab:bandwidth}
    \vspace{-3mm}
\end{table}

Table~\ref{tab:bandwidth} decomposes the CoCam4D message budget. Using only DOPs, we achieve 52.8 AP at 7\,KB/s, already outperforming CoCa3D at 1240\,KB/s, showing that object-level, velocity-aware DOPs carry substantially more task-relevant information per byte than BEV features.
Compared to VOGS-CP, which also transmits Gaussian primitives, CoCam4D attains a higher AP/KB ratio due to its object-centric design that avoids transmitting free-space Gaussians dominating outdoor scenes.
The full system, combining cached static Gaussians with per-frame DOPs, reaches 57.3 AP at 58\,KB/s, yielding an AP/KB ratio of 0.988 versus 0.041 for CoCa3D, i.e., a 24$\times$ improvement in communication efficiency.
\vspace{-20 pt}

\section{Conclusion}
In this work, we introduced \textbf{CoCam4D}, a cooperative camera-only 3D perception framework that addresses monocular geometry limitations via principled uncertainty modeling and fusion. By representing scenes as 3D Gaussian primitives with explicit covariances, it supports Bayesian multi-agent fusion where complementary views reduce depth uncertainty. Unlike heuristic feature sharing methods, it performs information-filter based integration in a single forward pass without LiDAR supervision or iterative optimization.  
We also proposed Dynamic Object Primitives (DOPs), a compact 4D representation for efficient uncertainty-aware communication under bandwidth constraints. Our analysis reveals a cooperative regime where camera-only systems match or exceed single-vehicle LiDAR performance.  

Overall, CoCam4D suggests a geometrically grounded and communication efficient direction for cooperative perception. As future work, it highlights uncertainty aware collaboration as a key step toward LiDAR free autonomous driving and toward joint perception and planning systems.

\bibliography{refer}
\end{document}